\documentclass[twocolumn]{article}
\usepackage{amsmath,amssymb,amscd,amsfonts,amsthm}
\usepackage{comment,mathtools}
\usepackage{url}
\usepackage{graphicx, xcolor}
\usepackage{paralist} 
\usepackage{bm}
\usepackage{booktabs}
\usepackage[round]{natbib} 
\usepackage{geometry}

\usepackage{authblk}

\usepackage{pifont}

\theoremstyle{definition}

\author[1]{Naofumi Hama}
\affil[1]{Hitachi, Ltd.\\Research \& Development Group}
\affil[ ]{\textit {naofumi.hama.hd@hitachi.com}}
\date{}
\title{Notes on Applicability of Explainable AI Methods to Machine Learning Models Using Features Extracted by Persistent Homology}
\begin{document}
\twocolumn[
    \maketitle
    \begin{abstract}
        Data analysis that uses the output of topological data analysis as input for machine learning algorithms has been the subject of extensive research.
        This approach offers a means of capturing the global structure of data. Persistent homology (PH),
        a common methodology within the field of TDA, has found wide-ranging applications in machine learning.
        One of the key reasons for the success of the PH-ML pipeline lies in the deterministic nature of feature extraction conducted through PH.
        The ability to achieve satisfactory levels of accuracy with relatively simple downstream machine learning models,
        when processing these extracted features, underlines the pipeline's superior interpretability.
        However, it must be noted that this interpretation has encountered issues.
        Specifically, it fails to accurately reflect the feasible parameter region in the data generation process,
        and the physical or chemical constraints that restrict this process.
        Against this backdrop, we explore the potential application of explainable AI methodologies to this PH-ML pipeline.
        We apply this approach to the specific problem of predicting gas adsorption in metal-organic frameworks and demonstrate that it can yield suggestive results.
        The codes to reproduce our results are available at \url{https://github.com/naofumihama/xai_ph_ml}.
    \end{abstract}
    \newline
]

\section{Introduction}
Typical topological data analysis (TDA) is used to examine the global structure of the data manifold that generates the actual data distribution scattered in the feature space.
The global properties of a manifold are generally defined by topological quantities that are invariant under continuous transformations.
For example, the number of holes in a manifold and number of its irreducible components, which are typically defined using the terminology of homology groups.
When applying such methods of algebraic geometry based on featureless empirical data points in the feature space,
it is convenient to consider a finite size around each data point and define the target manifold using their combinations.
The method of persistent homology (PH) characterizes the distribution of real data depending on
what type of global structure appears in accordance with its finite radius.

Data analysis using PH is not only used to extract properties of data manifolds from data distributions,
but also when spatially scattered point clouds form each data with a global structure, or when the organized information contained in the data is important.
PH extracts such information and characterizes each data.
Typically, the distribution of defects in material substances is characterized on the basis of the structure in accordance with their distribution,
and such characterization is used to estimate material properties.
A pipeline that uses the features extracted from PH as input to downstream machine learning models (PH-ML) has been gaining attention.
Since feature extraction by PH efficiently compresses meaningful parts of the structure within the data,
even with simple architectures for downstream ML, many problems can be solved with sufficient accuracy,
and it is also applied to image classification. Some of these applications are introduced in the next chapter.

In this pipeline, the feature extraction process mentioned above by PH is deterministic and intuitively interpretable.
Among the point clouds in the data, it is possible to specify with existing methods
which point becomes a boundary cycle that contributes significantly to downstream machine learning.
However, it is difficult to quantitatively evaluate how much each point in the point cloud contributed to the formation of the cycle.
Therefore, we consider whether it is possible to fairly evaluate these contributions by applying explainable AI (XAI) methods that are based on cooperative game theory.

As is often the case with the generation process of point clouds, when each data is characterized and generated from a smaller amount of parameters,
it is possible to evaluate the degree of contribution to the output values of downstream machine learning for each parameter
by applying methods classified as so-called black-box XAI to the entire pipeline.
However, there were concerns that this method, which obscures the mechanism of intermediate processing, might compromise the inherent interpretability of PH-ML.
We demonstrate that by evaluating the feature extraction process by PH from the high-order terms of the contributions calculated by applying XAI multiple times,
the specified parameters contribute to which cycle in PH and how they contribute in downstream machine learning can be visualized.
This method, while taking operable parameters as units, is believed to be able to directly contribute to improving experimental results
by providing information reflecting the mechanism of PH-ML.

In data like what we are considering in this letter, where the global structure is a problem, practical caution is required for the XAI method to be applied.
In point clouds that form a global structure, the distribution within the cloud is often subject to physical constraints.
When appropriately determining the contribution values from each feature while reflecting such constraints,
it is necessary to use an observational XAI method, rather than the intervention-type method often used as XAI, such as Kernel SHAP \citep{lundberg2017unified}.
We were able to extract more realistic results reflecting scientific properties by solving problems related to
the application of observational feature attributional XAI.

The rest of this letter is organized as follows.
In Section \ref{sec:rel}, we give examples of how the PH-ML pipeline is applied to real problems
and discuss XAI methods used in this letter.
In Sections \ref{sec:variable_len} and \ref{sec:higher_order}, we detail the problem of estimating gas adsorption in metal-organic frameworks (MOFs),
which is the problem setting we are specifically considering for possible application, and formalize our proposed method.
In Section \ref{sec:exp}, we apply our proposed methods to the problem of predicting MOF gas adsorption, showing that each produces meaningful results.
We conclude the letter in Section \ref{sec:conc}. The detailed setup of the experiment is presented in the Appendix.

\section{Related Works}
\label{sec:rel}
\subsection{Application Examples of Inputting Features Extracted from Persistent Homology into Machine Learning}
In this section, we give several examples of the PH-ML pipeline.
For a comprehensive understanding of the formation of PH,
its robustness based on the stability theorem \citep{cohen2005stability},
and its aptness for data analysis, the reader is referred to review literature,
for instance, see \cite{hensel2021survey, chazal2021introduction, obayashi2022persistent, ali2022survey}.
These studies provide an overview of the general properties of data analysis using PH.
\citet{hoef2022primer} also provide a meaningful review, covering from the basic definitions of PH to its applications in machine learning.
They demonstrated an example of classifying cloud satellite images on the basis of textual information.

The types of data represented by PH and their application have become diverse.
For instance, in condensed matter physics,
there are methodologies for estimating phases by inputting lattice spin configurations from the Ising model \citet{cole2021quantitative}
as well as classifying experimental result images \citet{leykam2022dark}.
In instances where input data are presented as waveforms and characterized using PH,
\citet{chung2021persistent} successfully classified whether the electrocardiogram waveform corresponds to wakefulness,
rapid eye movement (REM) sleep, or non-REM sleep.

There have been several studies on XAI using TDA.
These include efforts to characterize the feature space that the model processes via TDA, thereby acquiring information about the model.
For example, \cite{corneanu2020computing} aimed to characterize the data space and estimate the model's generalization ability.
\cite{xenopoulos2022topological} represented the contribution value vectors for each feature obtained through XAI using Mapper.

We applied PH-ML to the problem of estimating gas adsorption volume from the atomic configuration of MOFs.
\citet{townsend2020representation} used PH for feature extraction.
to address the problem of identifying molecules that selectively interact with carbon dioxide.

\subsection{Observational explainable AI}
\label{sec:obs_xai}
The process of data generation in the natural world is often constrained by scientific laws and other principles.
In the following sections, we consider the construction of machine learning models from such data to gain insights
into these scientific laws underlying data generation or exploring new data with desirable properties.
We also consider the use of XAI methods to elucidate the behavior of these models.
A widely used XAI method involves quantifying the contribution of each data feature to the model's output,
which is known as a feature attribution value.
A set of methods known as interventional XAI manipulate certain number of data features, input the altered data into the model,
and derive the feature attributions on the basis of on the difference between the output of the altered and original data.

This sensitivity analysis-like approach is among the best-known in XAI,
with the Shapley Additive Explanations (SHAP) \citep{lundberg2017unified} being particularly successful in computing desirable values
by distributing output differences among features based on cooperative game theory.
While interventional XAI methods are well-suited to obtain information about the behavior of the model,
they have been criticized for not necessarily being appropriate for gaining insights into the data \citep{chen2020true}.
This is primarily because they disregard the data generation process when altering feature values.

In contrast with interventional XAI, 
observational XAI takes into account the data generation process to obtain feature attributions.
This involves generating plausible data to compare with the target data,
or seeking characteristic properties of the target data using only actual data contained within the dataset.
These XAI methods are summarized in excellent surveys \citep{chen2023algorithms, olsen2023comparative}.
We consider the application of observational XAI to the PH-ML pipeline,
as it is deemed suitable for detecting insights from data by incorporating information related to the data generation process.

Cohort Shapley (CS) \citep{mase2019explaining} is a representative observational XAI method.
This method involves computing the conditional expectations appearing in the characteristic values of Shapley values
from the average of the output values assigned to a subset of the entire dataset.
It estimates the characteristic values when fixing specific features
by extracting data with similar values for those features in the target data and computing the average of the corresponding output values.
This approach is distinct in its capacity to derive the feature importance of each attribute of the target data exclusively from the supplied dataset,
without the necessity to specify any model as the explanation target.

An analogous method is integrated gradients cohort Shapley (IGCS) \citep{hama2022model}.
This method facilitates the application of integrated gradients \citep{sundararajan2017axiomatic},
founded on the Aumann-Shapley axioms, within the indicator space that delineates similarity evaluations in CS.
It achieves this by expanding it to continuous values via multilinear interpolation, rendering it differentiable.
It calculates the values of feature importance with similarly to CS in a higher-dimensional feature space and realistic computation time.

We applied CS and IGCS to a PH-ML pipeline developed on the basis of MOFs,
contingent on the number of dimensions of the features under consideration.
This enables us to quantify the contribution of each atom, or space not containing atoms, to the gas adsorption, which is the output of the PH-ML pipeline.
By calculating the contribution values from the persistent diagrams (PD) for a small number of building blocks that identify MOFs,
it is also possible to deepen the understanding of the entire pipeline.

Another distinctive method that should be added to the explainability for the PH-ML pipeline is the volume optimal cycle \citep{obayashi2018volume}.
This method identifies the feature points that form the boundary of a corresponding cycle when the birth and death are specified.
By applying regular XAI to the downstream ML model and discovering influential cycles with high contributions
then further applying this method, it is possible to allocate the contributions of the PH-ML pipeline to the feature points constructing the boundaries.
As illustrated in the example below, even if the feature points form the same boundary,
the importance of their formation should be different,
and this difference cannot be ascertained through a simple application of the volume optimal cycle.

\subsection{Gas Adsorption Estimation of MOF by PH-ML}
\label{sec:mof_setup}

In this section, we introduce a specific usecase of the PH-ML pipeline and discuss
applying a current XAI method to reveal what can be understood with existing method and its limitations.

The issue we discuss below is the estimation of gas adsorption in MOFs.
MOFs are composites of metals and non-metals,
and due to the large degrees of freedom in their structure,
they are being considered for designing new materials.
One application is selective gas adsorption.
By designing MOFs to have spaces within their crystal structure that can capture specific gas molecules,
gases can be selectively adsorbed.
For example, industrial applications are expected for new materials that can adsorb and immobilize greenhouse gases, such as carbon dioxide.

Simulations based on density functional theory can accurately determine the capacity to which each material adsorbs a specific gas at a fixed pressure,
but this simulation incurs high computational costs.
The design freedom of MOFs is vast,
and the exploration area for new materials is very wide,
so it is not realistic to give simulation-based evaluations to all candidates.
Therefore, primary filtering as a proxy for strict evaluation is necessary.
A strong alternative for this is evaluation by machine learning.
\cite{krishnapriyan2021machine} showed that the features obtained from the PH of the atomic coordinates of MOFs
are useful in downstream machine learning for estimating gas adsorption.
They estimated the adsorption of carbon dioxide and methane with high accuracy
using random forests with histograms of 1 and 2-dimensional PD landscapes as features.

We applied their framework to the Topologically-Based Crystal Constructor (ToBaCCo) dataset \citep{colon2017topologically}.
The MOFs included in this dataset are specified by several types of building blocks and templates specifying their relative positions.
A notable feature is that the scripts for reproducing the atomic configuration of MOFs when specifying these building blocks and a template can be publicly obtained.
In ToBaCCo dataset ver 1.0,
the methane adsorption based on simulation is annotated \citep{bobbitt2023mofx},
and each atom is specified by one type of template, up to two types of nodes, and up to one type of edge.
We refer to these up to four types of categorical parameters as "manipulatable parameters".
From this perspective, it is also possible to construct a pipeline as a regression problem that estimates gas adsorption by using the manipulatable parameters,
rather than the atomic coordinates themselves, as input.
When specifying operable parameters,
the atomic configuration of the unit cell of the crystal is obtained and is deprecated to fill a fixed size space ($100$ cubic angstrom in following) called a supercell.
The PH for the atomic coordinates contained in that supercell is calculated and input to the downstream machine learning.
Details on the configuration of PH and the construction of the random forest model are summarized in the Appendix.

\begin{figure}[htbp]
    \includegraphics[width=\linewidth]{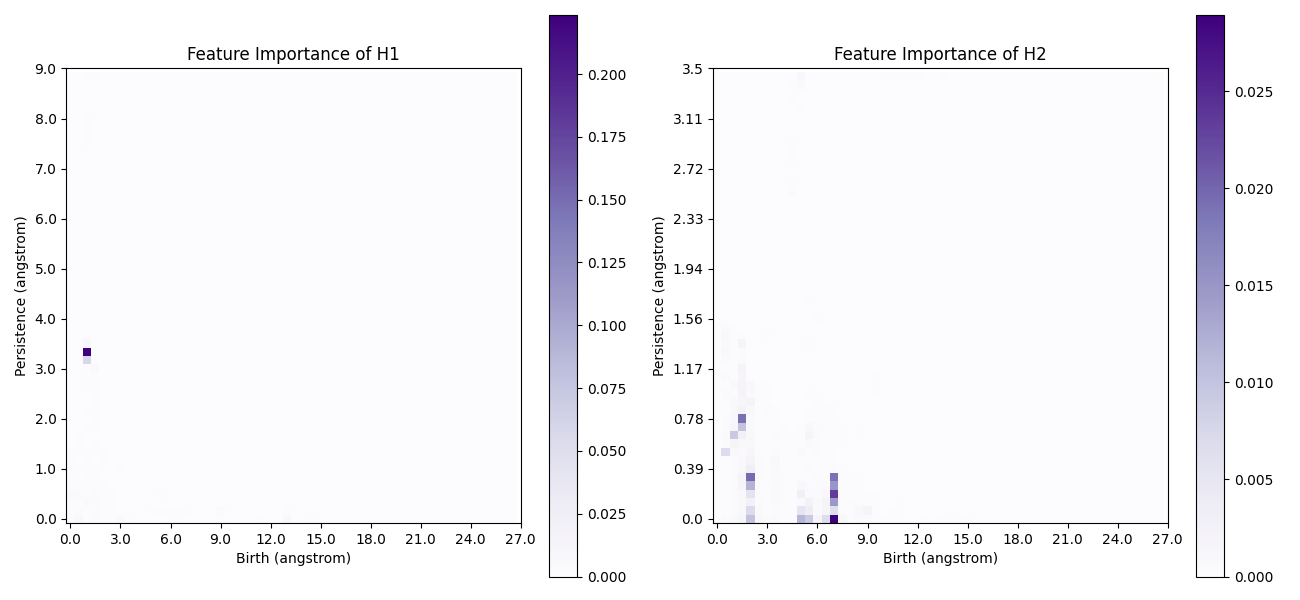}
    \caption{Permutation ordering of random forest model}
    \label{fig:rf_fa_global}
\end{figure}

The resulting random forest model achieved a $R^2$ coefficient of determination $0.861$ on the holdout data of ToBaCCo ver 1.0,
indicating that this pipeline is also effective for estimating gas adsorption in this dataset.
Permutation ordering was applied to this model,
and a heatmap of the importance of each pixel of the PD used as input was obtained, as shown in Figure~\ref{fig:rf_fa_global}.
Although a simple comparison is not possible due to the different datasets used,
the results do not match the interpretation associated with the typical size of methane ($3.8$ \AA) \cite{krishnapriyan2021machine}.

\section{Treatment as Variable Length Data}
\label{sec:variable_len}
\subsection{Problem Setting}
\begin{figure}[ht]
    \vskip 0.2in
    \begin{center}
        \centerline{\includegraphics[width=\columnwidth]{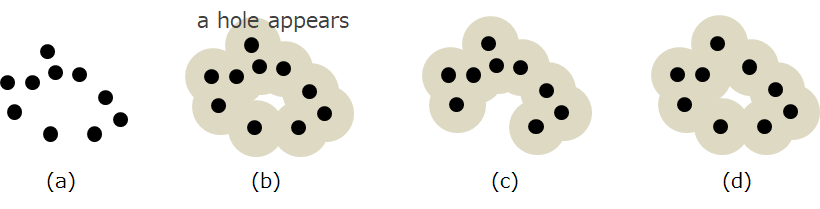}}
        \caption{Feature points in (a) are scattered spatially,
            and a hole appears in the certain radius ((b)).
            If black dot in bottom was not there ((c)),
            hole would not appear,
            but dot in dense area can be erased without changing homology ((d)).
            Therefore, there must be points that are more influential than others,
            and they should influential features to specify structure.
        }
        \label{fig:radius_distr}
    \end{center}
    \vskip -0.2in
\end{figure}

As mentioned above, by applying existing XAI methods to the PH-ML pipeline,
a certain degree of interpretability can be provided.
The feature extraction part of PH is a deterministic process and traceable.
However, we show that by quantifying the contribution from each feature point constituting the cycle of PH,
even higher interpretability can be provided.
This means that even if the influence from a specific cycle is required in the downstream machine learning,
each feature point constituting that cycle will not contribute equally to its birth and death.
An intuitive example is shown in Figure~\ref{fig:radius_distr}.

In this example,
for simplicity, we show the case of removing a single data point.
When such interventions are allowed, interventional XAI can be applied,
and by inputting the PH calculated before and after the removal of the coordination into the downstream ML,
the influence of that data point can be quantitatively assessed from the difference in outputs.
However, such interventions are not allowed in many data generation processes.
For example, it is impossible to remove a single atom from the inside of a crystal structure without compromising the overall stability.
Observational XAI, however, calculates feature importance only from existing data without using unnatural data intervened
(in terms of atomic coordination, for instance).
We consider the application of IGCS
as an observational XAI method that can be applied even when the feature space is high-dimensional.
As mentioned in Section \ref{sec:obs_xai},
IGCS compares the target data and the data within the population for each feature, which is a unit of explanation,
and calculates the feature importance based on whether they are similar.
Therefore, to apply IGCS, it is necessary to normalize so that the similarity can be defined for each feature as fixed-length data.
In atomic coordination of MOFs, the number and type of atoms vary for each MOF, which is variable-length data; and the order is not fixed;
and this normalization is non-trivial.
We propose a method called Grid-based Explanation to define this normalization.

\subsection{Grid-Based Explanation}
\label{sec:gbe}

Grid-based Explanation first sets a unit space large enough to include all feature points (atoms) within all data
then divides it into small grids on Cartesian coordinates.
Thus, the number of grids becomes common in all data.
In each data, the grid is then characterized by the number of feature points included in that grid.
Therefore, each grid characterized by the number of points included in it can be considered as a feature,
and the observational XAI method (IGCS for our case) can be applied to the output of the downstream machine learning.
Hence, the feature attributions allocated to each grid are evenly distributed to the feature points contained within,
so that the contribution value from each atom of the original data in the downstream ML can be calculated.

Applying observational XAI on the basis of this grid will result in distributing non-zero contribution values even to regions that do not contain any points.
This represents an evaluation of the impact that the absence of atoms in that spatial region has on the downstream ML.
Whether a region can accommodate gas molecules, for example, is often of significant importance in the PH-ML pipeline,
so this type of evaluation can have practical value. This is one of the advantages of applying observational XAI by using Grid-based Explanation.

By naturally normalizing the number of features of the data to the number of grids,
we make it possible to apply observational XAI.
When actually applying this Grid-based Explanation,
there are several points to address,
as discussed in the following sub-sections.

\subsubsection{Subprocess to Refer Spatial Symmetry}
\label(sec:rotsym)
In atomic configurations of crystals where feature extraction by PH is effective,
the absolute coordinates of each atom do not generally have a specific meaning, and only their relative coordinates matter.
In other words, these atomic coordinates often have translational and rotational symmetries.
When applying Grid-based Explanation to such systems, the divided grids should also reflect these spatial symmetries.
To do that, rotation and translation augmentations can be applied to the original dataset,
and the results can be included in the dataset targeted with IGCS.
This is one way to reflect the symmetry with Grid-based Explanation.
Therefore, before applying Grid-based Explanation,
it is necessary to refer to the distribution of the number of feature points contained in the grid and determine whether the system has spatial symmetry.

\subsubsection{Validity of Divides}
With Grid-based Explanation,
the contribution values allocated to each grid are evenly distributed to the feature points contained in the grid.
It should be added that this operation is justified by the Aumann-Shapley axioms to which IGCS conforms in a certain limit.
Assume that the size of the grid is sufficiently small compared with the birth and death of cycles extracted by PH.
From this assumption, once a point is included in the grid, the location and number of points in the grid do not affect the construction of the cycle.
Therefore, from the symmetry between the feature points inside the grid,
it can be seen that the contribution values assigned to the grid should be equally distributed to these feature points.

\section{Evaluation of Manipulatable Parameters to Influential Cycles}
\label{sec:higher_order}
\subsection{Problem Setting}
It is a natural assumption that data characterized by PH has their generation process controlled by a (relatively small number of) parameters.
For example, as mentioned in Section \ref{sec:mof_setup},
the MOFs included in ToBaCCo dataset ver 1.0 are uniquely determined by specifying only one template and several types of building blocks as parameters.
It is natural to capture the process of generating MOFs in terms of these manipulatable parameters
and determine the contribution of each manipulatable parameter to the downstream machine learning using XAI.

However, it is difficult to say that simply applying observational XAI to manipulatable parameters fully use the inherent interpretability that PH-ML possesses.
Observational XAI calculates on the basis of only the input and output of data without taking into account the data processing within the model,
so it cannot refer to the data generation process controlled by the manipulatable parameters, or the feature extraction process from data by PH.
Feature extraction by PH is a deterministic process,
highly interpretable in terms of reproducibility and traceability, and retains much information about its inner workings.
If we can refer to this, useful information could be obtained.
We demonstrated that by referring to this internal information,
we can obtain unprecedented granularity of information on how much each of the few manipulatable parameters contribute
to the composition of each cycle in the PD,
and how much that contributes to the downstream machine learning model.
In this context, the higher-order terms in conventional feature attributional XAI, usually interpreted as interaction terms, will provide this information.

\subsection{Higher Order Term Evaluation for Feature Engineering}
We first identify the contribution of cycles, represented as pixels in the landscape, to downstream ML using IGCS \citep{hama2022model}.
We then decompose each contribution assigned to these pixels into contributions from each manipulatable parameter using CS \citep{mase2019explaining}.
As a result of this operation, multiple heatmaps of the same size as the landscape can be drawn, corresponding to the number of manipulatable parameters.

Applying feature attributional XAI method multiple times to obtain higher-order terms is
usually used to determine the contributions from interactions between features \citep{janizek2021explaining,bordt2022shapley}.
However, our operation is not about higher-order terms among coequal features but between pixels of the persistent landscape and manipulatable parameters,
which are parameters of different hierarchies.
Considering the process in which data are generated from manipulatable parameters and a persistent landscape is constructed,
these higher-order terms can be perceived as quantities representing transitive relationships.

\section{Experiments}
\label{sec:exp}
We present the results of applying the proposed methods discussed in the previous chapters
to the problem of gas adsorption in MOFs in ToBaCCo dataset ver 1.0,
demonstrating their feasibility.

\subsection{Grid-based Explanation}
The MOFs generated using ToBaCCo dataset ver 1.0,
when given manipulatable parameters, assemble these parameters into building blocks that form a crystal structure,
filling the specified size as a supercell.
The generated MOFs are defined by the combination of the types of atoms and their absolute coordinates within the supercell.
The supercell does not necessarily have to be isotropic or limited to the shape of a cube, and the number of atoms is not constant.

We ignored the differences in the types of atoms and considered the process of extracting features only from the coordinates of atoms scattered within the supercell using PH.
Since the absolute coordinates within the supercell are specified in this manner,
we believe that the translational and rotational symmetries addressed in Section \ref{sec:gbe}
are compromised in the system under consideration.
However, we started by quantitatively verifying this.

If a building block is tilted, the absolute coordinate values can become negative.
After translating them to ensure they are zero or positive,
however, a cubic unit cell with a volume of $114$ cubic angstrom is prepared.
This is done to ensure it can contain building blocks of all types of MOFs.
This cube is divided into grids of 2 cubic angstrom each.
When the aforementioned absolute coordinates are applied to this,
the number of atoms contained in each grid can be used as a feature, enabling comparison of features across all MOFs.
Out of the all types of MOFs in the dataset,
200 were randomly selected and characterized based on the number of atoms in the $57^3=185,193$ grids thus created.
Out of these grids, $155,881$ did not contain any atoms from any MOF, making them meaningless as features.
Even for the other grids, the distribution of which a MOF contains atoms in which the grid is heavily skewed.
This shows that the dataset in question does not have translational symmetry with respect to spatial coordinates,
so the augmentation discussed in Section~\ref{sec:gbe} is unnecessary.

\begin{figure}[ht]
    \vskip 0.2in
    \begin{center}
        \centerline{\includegraphics[width=\columnwidth]{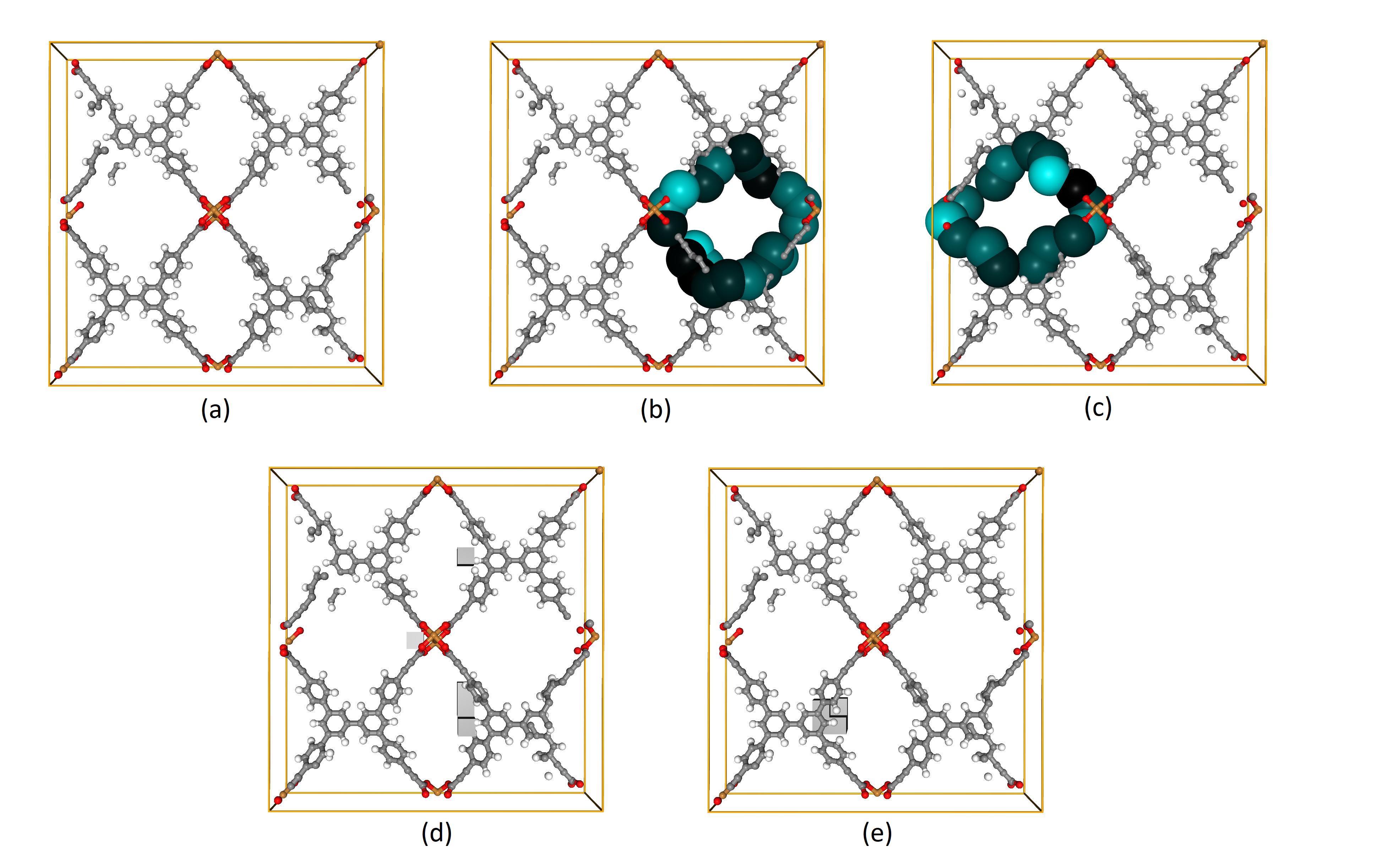}}
        \caption{(a): Unit cell of the molecule drawn by \cite{rose2015ngl}.
            (b): Contribution from atoms that compose boundary of
            $(birth, death)\sim(2.42,5.93)$.
            Brighter atoms have large positive contribution
            .
            The radii of atoms correspond to the birth time of the cycle.
            (c): Contribution from atoms that  are boundary of
            $(birth, death)\sim(2.48,5.55)$.
            (d): Five largest positive regions without atoms.
            (e): Five largest negative regions without atoms.}
        \label{fig:unit_vanilla_4921}
    \end{center}
    \vskip -0.2in
\end{figure}

\begin{figure}[ht]
    \vskip 0.2in
    \begin{center}
        \centerline{\includegraphics[width=\columnwidth]{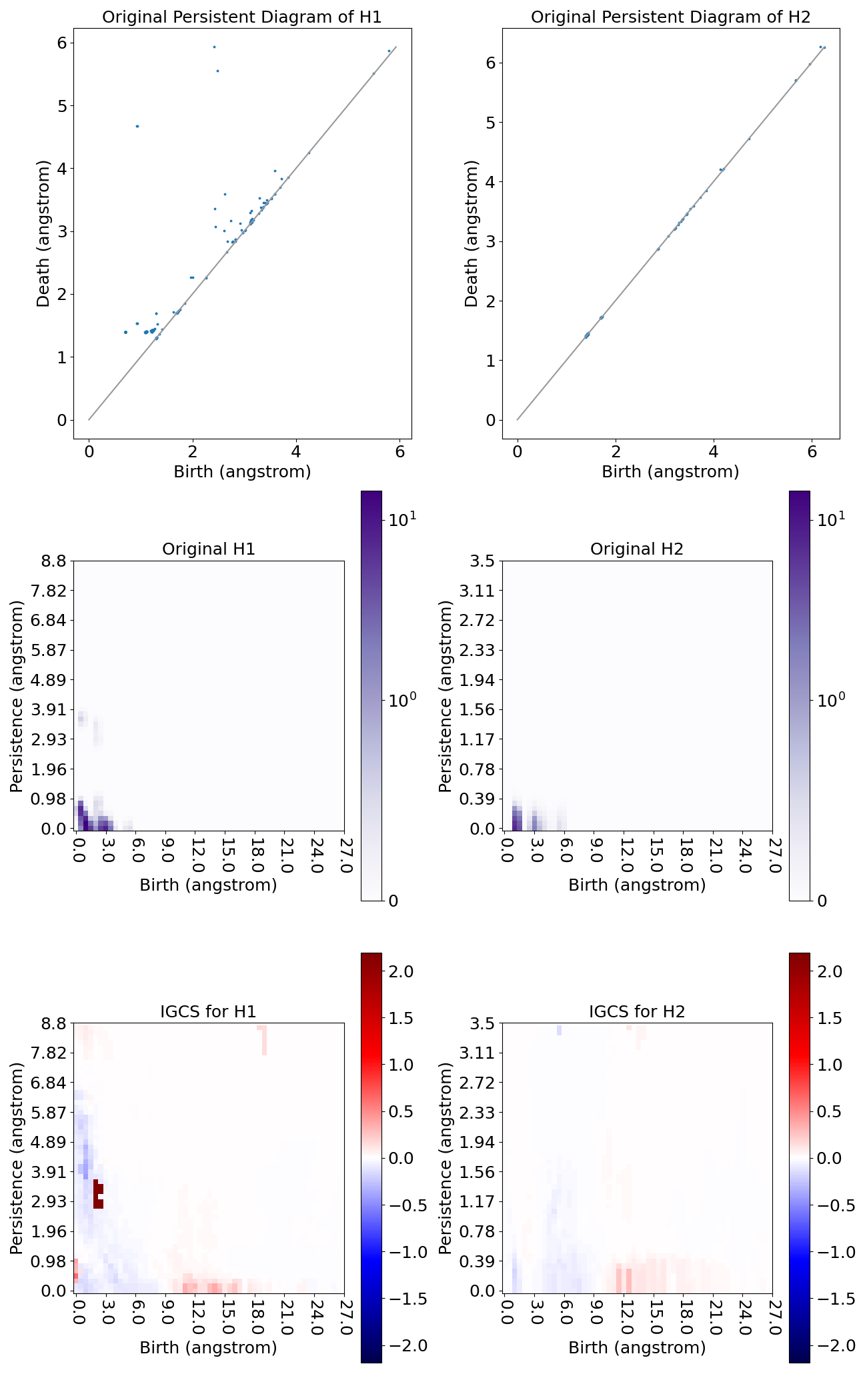}}
        \caption{PDs of $H_1$ and $H_2$ of molecule in Figure~\ref{fig:unit_vanilla_4921} (a) (top graphs),
            corresponding landscapes (middle graphs)
            and heatmap of the model drawn using IGCS (bottom graphs)
        }
        \label{fig:tobmof_4921}
    \end{center}
    \vskip -0.2in
\end{figure}

One MOF from the dataset is specifically selected to exemplify the explanation obtained with Grid-based Explanation.
The selected MOF is identified by the manuplatable parameters such as
Template: lvtb, Node1: sym5\_on\_10, Node2: sym5\_mc\_2, Edge: L\_19.
The unit cell of the molecule configured by them is as shown in Figure~\ref{fig:unit_vanilla_4921} (a).
This molecule has a flat unit cell in the depth direction of the figure.
The gas adsorption level estimated with the downstream machine learning model for this molecule is 221.47 ($\mathrm{cm}^3 (\mathrm{STP}) / \mathrm{cm}^3 $).
The PDs for the first and second dimensions are shown in the top and middle graphs of Figure~\ref{fig:tobmof_4921}, respectively.
The values that each pixel of the landscape, drawn using IGCS comparing with the other 999 molecules of this dataset,
contributes to the gas adsorption level, as shown in the bottom graphs of Figure~\ref{fig:tobmof_4921}.
Most of the positive contributions to gas adsorption come from the cycles
located at $(birth, death)\sim(2.42,5.93)$ and $(birth, death)\sim(2.48,5.55)$ in $H_1$.

With Grid-based Explanation, it is necessary to define cohort data to apply observational XAI.
As mentioned in Section \ref{sec:obs_xai},
the main difference between observational XAI and interventional XAI is that
observational XAI is compromised as soon as data that ignore scientific constraints and cannot actually exist are used in the cohort.
In our approach, which characterizes MOFs only by the number of atoms contained in the grid,
simply using MOFs generated from other manipulatable parameters into the cohort and applying Grid-based Explanation makes them very different as comparison targets,
making it difficult to obtain useful information to find new and better material candidates.
Therefore, by making molecules that underwent structural relaxation after adding perturbations to the crystal structure of the MOF in question a cohort,
useful information can be extracted as a valid comparison target.
This perturbation involves moving all atoms in a random direction by a fixed length (1\AA).
The perturbation and structural relaxation are done using methods implemented in pymatgen \citep{ong2013python} with default settings.
Whether this application is scientifically permissible goes beyond the scope of this letter.

We included the 584 atomic coordinates obtained with this operation in the cohort and applied Grid-based Explanation.
Note that the grid size was set to $2$ cubic angstrom.
It was found that all grids containing one or more atoms of this molecule made a positive contribution.
This is because, compared to any comparison target in the cohort obtained by the perturbation, the molecule in question has a higher gas adsorption level.

Regarding the two influential cycles mentioned above that give a significant positive contribution in Figure~\ref{fig:tobmof_4921},
the contributions of each atom forming the boundaries of these cycles are visualized in Figure~\ref{fig:unit_vanilla_4921} (b) and (c).
These atoms were identified with the volume optimal cycle and are color-coded in accordance with to the contribution values calculated using Grid-based Explanation.
Due to perturbations in position,
we obtained the contribution that each atomic position has to the output values of the downstream machine learning,
from a robustness perspective, where changes in cycle size alter estimated gas adsorption levels, for instance.
As shown in Figure~\ref{fig:unit_vanilla_4921} (c),
the atom that contributes strongly positively (hydrogen atoms bonded to the aromatic ring) are adjacent to
the atom with a small contribution (carbon atoms connecting building blocks).
This suggests that, from a gas adsorption perspective, the latter atoms make a relatively weak contribution,
and if a user tries to replace wome of the building blocks in this molecule with a new type of building block to further improve the gas adsorption level,
keeping the position of the former hydrogen atom and searching for a building block that can only change the position of the latter carbon atom would be a promising guideline.

The gray boxes in Figure~\ref{fig:unit_vanilla_4921} (d) indicate empty grids that contain no atoms,
highlighting the top five grids making significant contributions.
This suggests that the absence of atoms in these areas enhance the gas adsorption.
There are grids adjacent to the atoms in Figure~\ref{fig:unit_vanilla_4921} (c) that make only a small contribution.
When searching for a new building block to improve gas adsorption,
users should choose blocks that will not position atoms in these gray areas.

Conversely, in Figure~\ref{fig:unit_vanilla_4921} (e), the gray boxes represent empty grids that have strong negative contributions.
It is interpreted that if these areas contained atoms,
gas adsorption could have been further improved.
This also provides important implications for improving gas adsorption with new building blocks.

Thus, Grid-based Explanation is applicable to the PH-ML pipeline,
offering a method capable of extracting practical information from the data.

\subsection{Higher Order Terms Evaluation for Feature Engineering}
\label{sec:higher_exp}
In this section, using ToBaCCo dataset ver 1.0 as an example again,
we demonstrate that the transition relationships, from manipulatable parameters to the features extracted by PH and the outputs of downstream machine learning,
can be quantitatively obtained through the higher-order terms of feature attributional XAI.

\begin{figure}[ht]
    \vskip 0.2in
    \begin{center}
        \centerline{\includegraphics[width=\columnwidth]{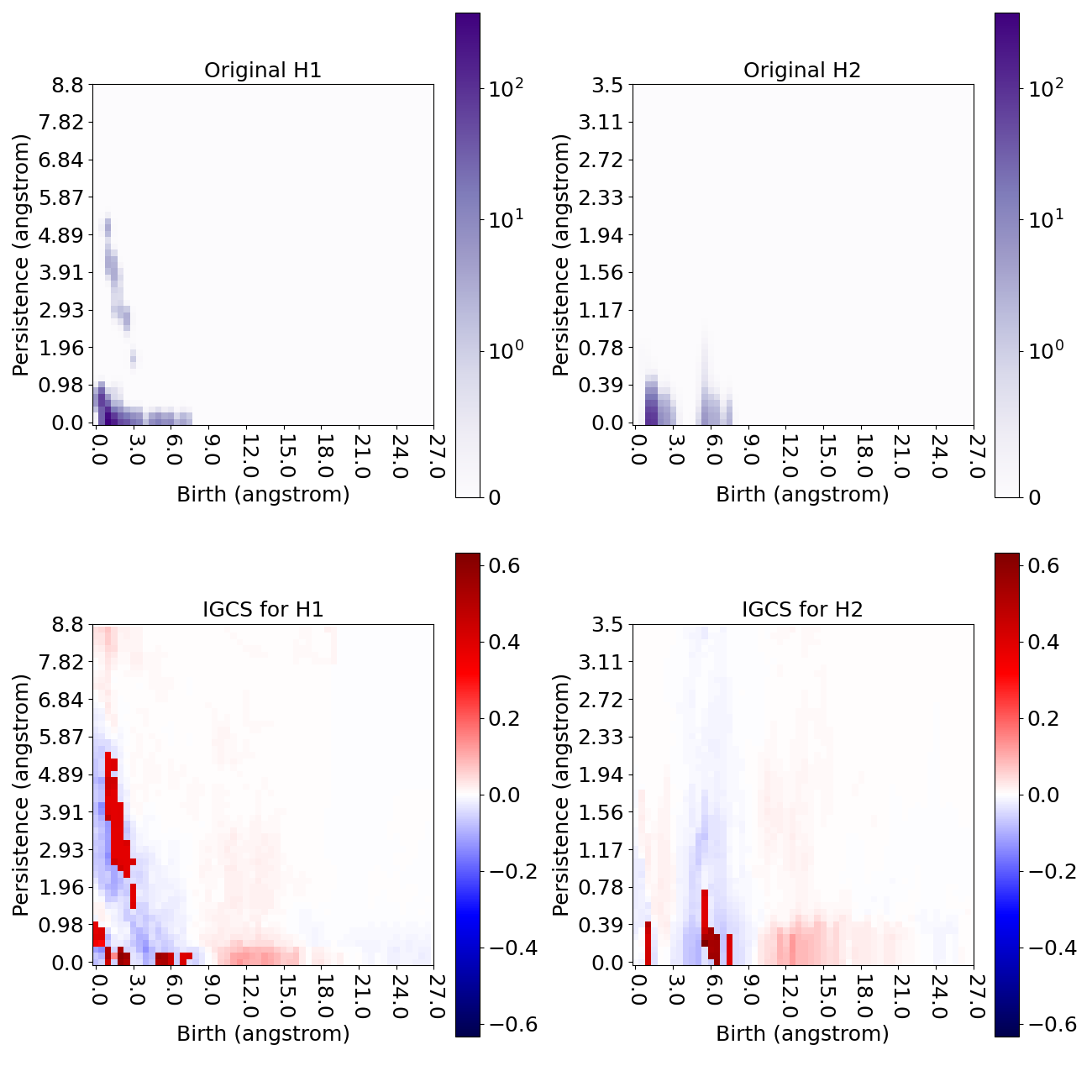}}
        \caption{XAI results from IGCS on MOF
            whis parameters Template: ith, Node1: sym4\_on\_14,
            Node2: sym13\_mc\_12, Edge: L\_43.
            Top graphs: landscapes from PH.
            Bottom graphs: attributions of each pixel by using IGCS.}
        \label{fig:tobmof_existing_4049}
    \end{center}
    \vskip -0.2in
\end{figure}

The MOF of interest for our explanation is specified by the parameters Template: ith, Node1: sym4\_on\_14, Node2: sym13\_mc\_12, Edge: L\_43.
This MOF has an annotated methane adsorption value of 257  ($\mathrm{cm}^3 (\mathrm{STP}) / \mathrm{cm}^3 $)  under 100 bars.
The inference value from the machine learning model we use is similarly $237.35$ ($\mathrm{cm}^3 (\mathrm{STP}) / \mathrm{cm}^3 $).
Figure~\ref{fig:tobmof_existing_4049} illustrates the contribution magnitude for each pixel of the landscape, calculated using IGCS.
From another 999 MOFs specified by other manipulatable parameters,
each of their persistent landscapes was used as a cohort.

IGCS can be applied directly to the relationship between the manipulatable parameters and gas adsorption.
Within the cohort of 1,000 data points including the target data,
there were 24 MOFs with the parameter edge: L\_43.
Applying IGCS, it is clear that in most cases parameter L\_43 hads a negative contribution for these 24 MOFs (average: $-13.45$),
but for our data of interest, it hads a significant positive contribution ($44.68$).

\begin{figure}[h]
    \vskip 0.2in
    \begin{center}
        \centerline{\includegraphics[width=\columnwidth]{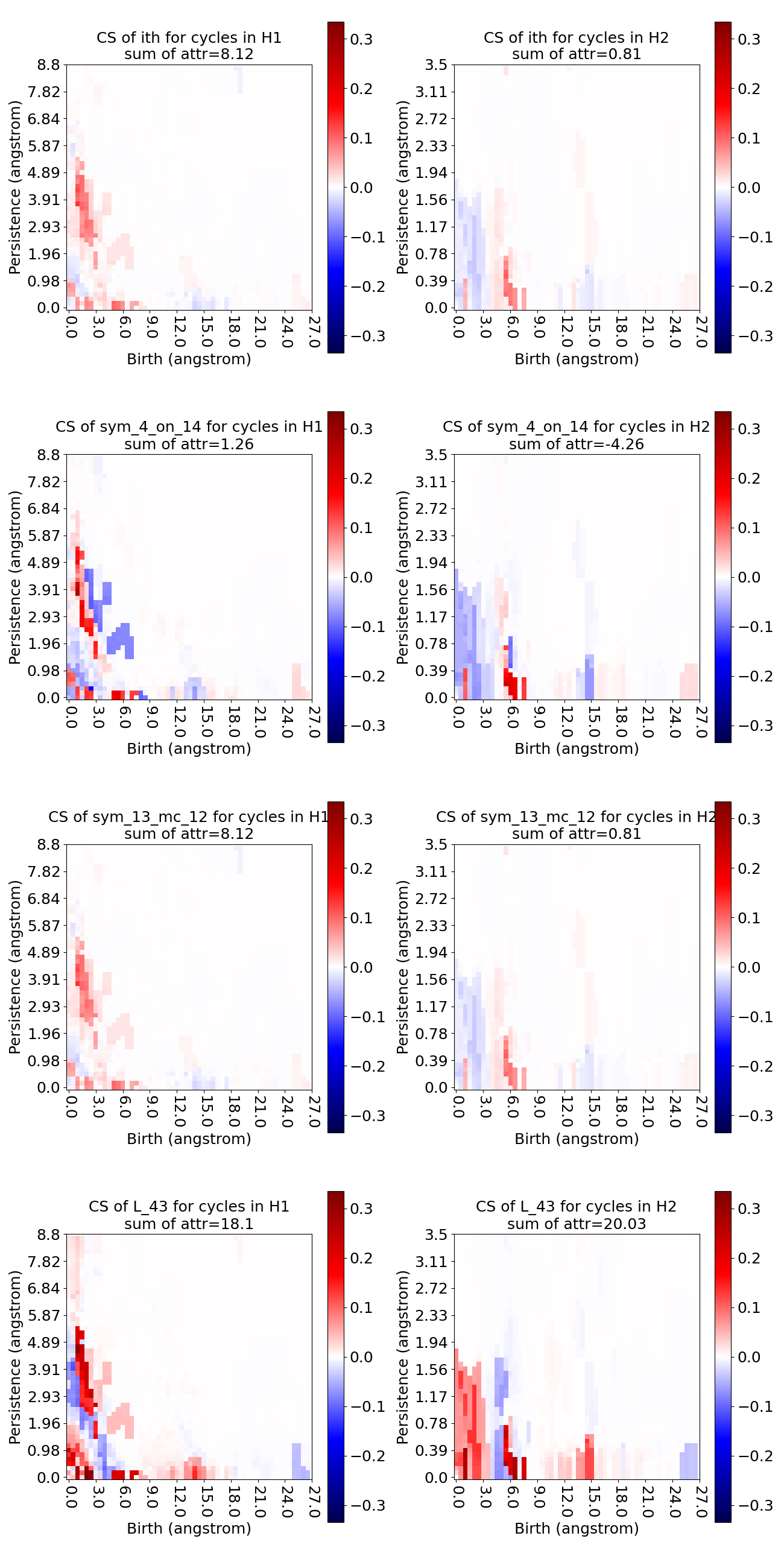}}
        \caption{Higher order terms evaluation of landscapes
            attributed from each building block}
        \label{fig:tobmof_cs_for_igcs_4049}
    \end{center}
    \vskip -0.2in
\end{figure}

The following question arises: why does L\_43 have a positive contribution in this molecule?
The answer can be found in the transitive relationship using the higher-order terms.
The contribution values shown in the heatmap in Figure~\ref{fig:tobmof_existing_4049}
are further decomposed into four types of manipulatable parameters by applying CS,
the results of which are shown in Figure~\ref{fig:tobmof_cs_for_igcs_4049}.

The absence in this molecule of short-lived cycles within $H_1$ and early-born cycles within $H_2$,
which have negative contributions to gas adsorption in other manipulatable parameters, is offset by the positive contributions of L\_43,
changing the contribution from each pixel to zero or a weak positive value for the whole molecule.
Conversely, if the edge parameter is changed from L\_43 to another value,
it is anticipated that cycles corresponding to these pixels will emerge, contributing negatively to gas adsorption.

One of the distinguishing features of this method using higher-order terms is the ability to glean information reminiscent of causal inference.
Users can refer to this information alongside domain knowledge, enabling them to assess the validity of data processing through the PH-ML pipeline.
By providing this higher level of interpretability, it is expected that users' trust in the PH-ML pipeline will be bolstered.

\section{Conclusion}
\label{sec:conc}
We explored the feasibility of applying feature attributional XAI to problems
where structural information from data is extracted using PH and used as features for machine learning.
We also specifically addressed the issue of estimating gas adsorption in MOFs and verified what useful information can be obtained.
Thus, we were able to propose more suitable methods of applying XAI when non-local information from data are used through PH.

\section*{Acknowledgement}
We thank Art B. Owen and Masayoshi Mase for for discussions. We also thank to anonymous reviewers to improve this letter.

\bibliographystyle{apalike}
\bibliography{xtda_bib}

\appendix

\section{Detailed Model Descriptions of the Experiments}\label{sec:exp_detail}
This appendix provides background on the experiments conducted for this letter.

\subsection{Random Forest Model}
We used a random forest regressor implemented in scikit-learn \citep{scikit-learn}.
The number of estimators was $500$.
Other hyperparameters were set as the default setting of the scikit-learn.
The heatmap in Figure~\ref{fig:rf_fa_global} is impurity-based feature importance,
also implemented in scikit-learn.

\subsection{Feature Extraction by PH}
Feature extraction by PH has several hyperparameters.
We used the softwares, Dionysus \citep{morozov2007dionysus} and Homcloud \citep{obayashi2022persistent},
to draw the PDs.
We put cutoffs to birth times and persistence lengths as summarized in Table~\ref{tb:cutoff_ph},
as maximum lengths.
The resulting PDs were discretized into two-dimensional histograms,
the number of bins were $54 \times 54$.
PDs were blurred in Gaussian with the standard deviation set to $0.15$.
The settings for these figures were determined after confirming through preliminary experiments that the downstream random forest model,
using the features summarized as histograms, could perform adequately.

\begin{table}[t]
    \caption{Maximum length of birth and persistence of cycles as cutoff.}
    \label{tb:cutoff_ph}
    \vskip 0.15in
    \begin{center}
        \begin{small}
            \begin{sc}
                \begin{tabular}{llr}
                    \toprule
                    Dimension & Parameter    & Maximum length \\
                    \midrule
                    $H_1$     & Birth        & 27.0           \\
                    $H_1$     & Perisistence & 8.8            \\
                    $H_2$     & Birth        & 27.0           \\
                    $H_2$     & Perisistence & 3.5            \\
                    \bottomrule
                \end{tabular}
            \end{sc}
        \end{small}
    \end{center}
    \vskip -0.1in
\end{table}

\subsection{Observational explainable AI}
We used CS \citep{mase2019explaining} and
IGCS \citep{hama2022model}
to obtain feature attributions from existing data.
CS and IGCS have several hyperparameters.
We reduced the number of steps of Riemann sum with IGCS to 50 (default:500) to reduce computational costs,
and set the ratio of similarity threshold to $0.01$ (default:$0.1$),
as pixels in landscapes take values in an exponential scale
and the default value is too rough to define similarities between them.
For categorical values in the parameters of ToBaCCo 1.0, the levels of which can be more than 40,
the thresholds of similarity for them, mentioned in Section~\ref{sec:higher_exp} is considered small enough
to distinguish them.

\end{document}